\documentclass[lettersize, journal, margin=2cm, pdftex]{acmart} %manuscript, screen, review \ifx\HCode\UnDef\else\hypersetup{tex4ht}\fi
\usepackage{amsfonts} %amssymb,
\usepackage{algorithmic}
\usepackage{algorithm}
\usepackage[caption=false,font=normalsize,labelfont=sf,textfont=sf]{subfig}
\usepackage{textcomp}
\usepackage{stfloats}
\usepackage{url}
\usepackage{verbatim}
\usepackage{graphics}
\usepackage{bmpsize}
\usepackage{natbib}
\usepackage{xcolor}
\usepackage{wrapfig}
\usepackage{booktabs}
\AtBeginDocument{%
  }

\begin{document}

\newcommand*{\change}[1]{\textcolor{black}{#1}}

\title{Music Mode: Transforming Robot Movement into Music Increases Likability and Perceived Intelligence}

\author{Catie Cuan}
\affiliation{%
  \institution{Stanford University}
  \streetaddress{440 Escondido Mall}
  \city{Stanford}
  \state{California}
  \country{United States}
}
\email{ccuan@stanford.edu}

\author{Emre Fisher}
\affiliation{%
  \institution{ASML}
  \streetaddress{125 Rio Robles}
  \city{San Jose}
  \state{California}
  \country{United States}}
\email{fisheremre@gmail.com}

\author{Allison Okamura}
\affiliation{%
  \institution{Stanford University}
  \streetaddress{440 Escondido Mall}
  \city{Stanford}
  \state{California}
  \country{United States}}
\email{aokamura@stanford.edu}

\author{Tom Engbersen}
\affiliation{%
  \institution{Skydio}
  \streetaddress{3000 Clearview Way}
  \city{San Mateo}
  \state{California}
  \country{United States}
}
\email{tomengbersen@gmail.com}

\renewcommand{\shortauthors}{Cuan et al.}

\begin{abstract}
  As robots enter everyday spaces like offices, the sounds they create affect how they are perceived. We present ``Music Mode,'' a novel mapping between a robot's joint motions and sounds, programmed by artists and engineers to make the robot generate music as it moves. Two experiments were designed to characterize the effect of this musical augmentation on human users. In the first experiment, a robot performed three tasks while playing three different sound mappings. Results showed that participants observing the robot perceived it as more safe, animate, intelligent, anthropomorphic, and likable when playing the Music Mode Orchestra software. To test whether the results of the first experiment were due to the Music Mode algorithm, rather than music alone, we conducted a second experiment. Here the robot performed the same three tasks, while a participant observed via video, but the Orchestra music was either linked to its movement or random. Participants rated the robots as more intelligent when the music was linked to the movement. Robots using Music Mode logged approximately two hundred hours of operation while navigating, wiping tables, and sorting trash, and bystander comments made during this operating time served as an embedded case study. \change{This paper has both designerly contributions and engineering contributions.} The contributions are: (1) an interdisciplinary choreographic, musical, and coding design process to develop a real-world robot sound feature, (2) a technical implementation for movement-based sound generation, and (3) two experiments and an embedded case study of robots running this feature during daily work activities that resulted in increased likeability and perceived intelligence of the robot. 
\end{abstract}





\maketitle

\section{Introduction}

Helper robots, a new category of robots that assist in everyday spaces like offices, grocery stores, hotels, restaurants, and homes, may increase humans' productivity and free time. These robots perform a medley of tasks, such as dish collection, item delivery, grocery aisle inspection, and table wiping. Many robots in this category are mobile and fully or mostly autonomous. They can appear in several form factors, from the column-like Badger\cite{BadgerRobot} to the boxy TUG \cite{TUGRobot}. 
\begin{wrapfigure}{r}{0.6\textwidth}
  \begin{center}
    \includegraphics[width=0.58\textwidth]{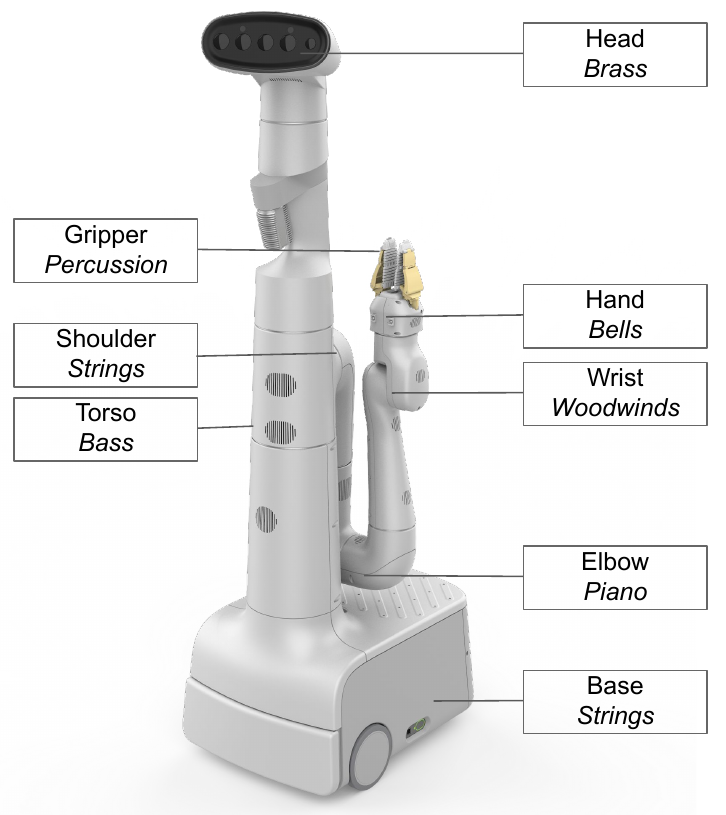}
  \end{center}
    \caption{Each label corresponds to a specific location on the robot, indicating the joint and corresponding musical instrument. For example, the ``Gripper'' corresponds to the ``Percussion'' sound. The robot is a prototype created by Everyday Robots.}
    \label{glamour}
\end{wrapfigure}For these robots to be welcomed in public spaces, where they will be moving around and helping a diverse population of humans, they should be designed with users' and bystanders' needs and preferences in mind.

Helper robot design teams may include social robotics researchers, animators, composers, psychologists, anthropologists, interaction designers, industrial designers, mechanical engineers, and software engineers. These different forms of expertise are necessary due to the robot's embedding in an everyday environment. Prior work has shown that artistic expertise, from theater to choreography, can inform the design of social robots to make them not only welcomed, but also trusted, friendly, and supportive \citep{cuan2018curtain, hoffman2014designing, jochum2016using, rond2019improv}.  

The robot described in this paper is a prototype robot from Everyday Robots that was deployed at Alphabet offices in Mountain View, California in 2021-2023. Over 100 of these robots worked in several buildings, where they perform tasks like wiping tables and sorting trash. These robots worked nearby hundreds of people per week who have a wide range of jobs, from software engineers to chefs. Because the robots were prototypes, they were monitored by a co-located human operator while the robots perform autonomous tasks. The Everyday Robots design team experimented with different robot interaction modes and communication signals as they, like other helper robot companies and researchers, pursued the aim of building an accepted, legible, and delightful robot for everyday spaces.

This paper describes a prolonged research and creative collaboration between engineers and the Artist-in-Residence at Everyday Robots and the resulting generative sound feature that arose from this collaboration. \change{As articulated in \citep{lupetti2021designerly}, this paper highlights both the conceptual contribution of this design process (which, like prior work in designerly HRI, included artists and subjectivity) as well as the artifact of the software and robot sounds generated.} The mapping between sounds and joints is shown in Figure \ref{glamour} and videos can be found in the supplementary material. The contributions of this work are threefold: (1) an interdisciplinary choreographic, musical, and coding design process to develop a real-world robot sound feature, (2) a technical implementation for movement-based sound generation, and (3) two experiments and an embedded case study of robots running this feature during daily work activities that show the resulting changes in robot perception. Section \ref{RelatedWork} describes relevant prior work in artistic robotics and robotic sounds in human-robot interaction. Section \ref{Design_Process} explains the artistic design process, technical implementation, and audio implementation. Section \ref{Methods} describes the experiment design. Section \ref{Results} details the outcome of the two controlled experiments as well as the on-site case study, followed by Section \ref{Discussion} which discusses future work.
\section{Related Work} \label{RelatedWork}

\subsection{Sound Design for Human-Robot Interaction}
Researchers have proposed sonic design principles for human-robot interaction based on specific goals, such as character or interactivity \citep{robinson2022designing}. For example, the aim of the SONAO project is to ``establish new methods based on sonification of expressive movements for achieving a robust interaction between users and humanoid robots'' \citep{bresin2021robust}. Many studies have measured the effects of different sounds on human perception of robots \citep{zhang2021bringing, latupeirissa2020exploring, holland2013music}. One study found that incorporating music with matching movement makes robots appear more human-like and increases positive feelings towards the robot \citep{hoffman2013effects}. In addition to the human perception benefits, sonic robotic expression has also been shown to assist in people localizing robots in space \citep{cha2018effects} and inferring clarity of vehicle interntion when interacting with autonomous cars \citep{moore2020sound}. Not all sounds lead to improved robot impressions; prior work found that machine sounds intrinsic to a robot, such as as servos rotating while the robot actuates, may be disliked overall \citep{moore2017making, moore2019unintended}. \change{Composers have been involved in not only several human-robot interaction projects but also designs of other technologies like autonomous vehicles \citep{misdariis2019electric}.}

\change{In \citep{robinson2021smooth}, researchers composed different sound conditions and combined them with videos of robot motion, a method of ``designing sound as an accompaniment to robot movement with the aim of conveying information.''} They found that different sound conditions affected perceived robot attractiveness, movement quality, and safety. Researchers have also examined how different sound waves produced by a robot can be perceived as different emotions depending on the shape, wavelength, and magnitude of the wave \citep{song2017expressing}. \change{In \citep{zahray2020robot}, researchers tested several different sonification techniques to map motion to pitch/timbre, emotion-based sounds, and velocity-based sounds. Researchers also created a series of gestures to convey emotions, and then sonified those gestures to improve the correct identification of the demonstrated emotion \citep{frid2022perceptual}.}

\change{Our work differs from these projects in that the sound conditions were generated from a collaborative artistic-engineering design process, were deployed on a real robot (through its on-board computer and speaker) with sounds cued from live sensor streams, and the experiments and case study were performed in situ while the robot performs its everyday tasks (rather than pairing a specific emotion with a sound). Further, our work included participants that had frequently worked with the robot, demonstrating its effects on short-exposure participant populations (exposed to the robot in a one-time user study) as well as long-exposure participant populations (previously exposed to the robot).}

\subsection{Robots Playing Musical Instruments and Robots as Music Instruments}
Researchers have taught existing robots to play regular and custom musical instruments and designed robots for the express purpose of playing instruments \citep{weinberg2006toward, kapur2005history}. Shimon, an interactive robotic marimba player, utilized an action framework in order to play with other human musicians \citep{hoffman2010shimon}, and software was developed to dictate the interaction modes between the robot player and humans \citep{weinberg2009interactive}. In prior work, humanoid robots were trained to play instruments designed for humans such as a keyboard \citep{kim2011enabling} or a Theramin \citep{mizumoto2009thereminist}.

Engineers, researchers, and musicians have also designed robots that function as musical instruments \citep{kapur2011machine, singer2004lemur, singer2005large, singer2003lemur, murphy2014expressive}. Novel software was developed in several such projects, such as software to control each aspect of the robotic musical instrument as an agent \citep{eigenfeldt2008agent} or software to create an interface that interpreted human commands into robot music \citep{suzuki2004robotic}. In contrast, the robots in our work were built to perform useful tasks and the Music Mode software was added as an expressive layer to augment the robots' regular behavior. Our work also included various music soundscapes and profiles that altered humans' overall impression of the robot. 

\change{Researchers have explored linking a robot's gestures to non-verbal prosody to enhance emotion recognition \citep{savery2019establishing}. Researchers have also explored enriching a music playing robot's voice with generative prosody, creating a more musical exchange between a music-playing robot and an improvising human before and after a performance. This study found that the prosody condition led to higher likeability and perceived intelligence ratings \citep{savery2021before}. In our work, we use exclusively generated music from a series of samples rather than musical prosody, and test this effect across functional tasks in different office settings, rather than linking the sound effects exclusively to emotion.} 

\subsection{Artists and Artwork in Human-Robot Interaction Design}
Theatrical and dance artists have collaborated with human-robot interaction researchers on numerous robot designs, experiments, and robot movement generation \citep{knight2011eight, rond2019improv, hoffman2014designing, sirkin2014using, laviers2018choreographic, cuan2018curtain, cuan2019time}. Musicians and composers have also collaborated with roboticists on several such projects. While making an interactive sonic robotic artwork, researchers treated sound-based human-robot interaction as a testbed for interplay between electroacoustic music and robot sound \citep{robinson2022crafting}. In \citep{dahl2017data}, researchers studied how trained musicians created sound for robot movement, which led to a sound synthesis application \citep{bellona2017empirically}. 

\begin{figure}[b!]
\centering
\includegraphics[width=\textwidth]{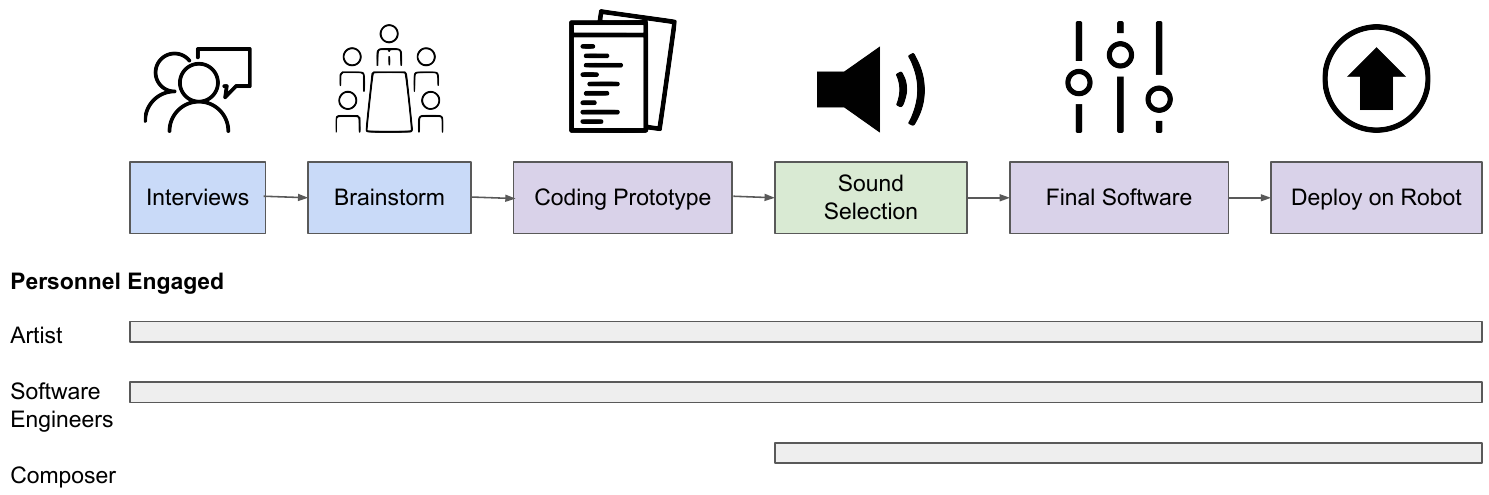}
\caption{The design process for the Music Mode feature. Interviews were followed by group Brainstorming. After the first Coding Prototype was generated, we did an initial Sound Selection. This led to the Final Software that was pushed onto all 180+ robots during the Deploy on Robot stage. The elements in blue are group verbal design methods, elements in purple involve coding, and elements in green are artistic efforts.}
\label{designprocess}
\end{figure}

Robots have appeared in live performances and art installations, where they generated sounds as part of their presence \citep{snyder2015machine, hoffman2011stage, katevas2014robot, szecsei2019theatrical}. For example, silent robots appeared alongside human-generated sound in an interactive artistic exhibit \citep{breazeal2003interactive}. Researchers studied the sounds generated by robots in films in order to inspire sonic source material for robot motion \citep{latupeirissa2019sonic}, and defined categories for sound design based on robot motion \citep{latupeirissa2020understanding}.

\section{Design Process} \label{Design_Process} %Rename design process?

\subsection{Initial Overview}

At the start of the residency in Winter 2021, the artist conducted a series of in-depth interviews with team members at Everyday Robots (``Interviews'' portion of Figure \ref{designprocess}) and on other teams in order to understand how they perceived the robot. Frequent descriptions included, ``clean'', ``simple'', ``useful'', and ``humble''. The artist aimed to elevate the robot from a utilitarian entity into a compelling and delightful agent, such that it would be welcomed in different spaces and inspire positive sentiments. 

The central goal was to experiment with interaction modes and robot qualities without altering the robot physically. This meant that the artist was able to change the robot's light, movement, and sound exclusively through software. A group of engineers and product managers who were curious about the artistic process reached out to the artist after the initial interviews, and this group began to meet with the artist every two weeks. During these meetings, the group discussed different artistic prompts and how these prompts may be realized with the robots and software platform. These meetings built rapport and trust between the artist and this group, in addition to clarifying the scope and scale of the work possible within the residency period.

After this group was formed, the team members engaged in frequent Brainstorming sessions (Figure \ref{designprocess}). A frequent theme explored in these meetings was robots becoming more acceptable and welcomed by generating or playing music. Ideas varied from robots playing large bells and distributed pipes to robots playing existing drums and stringed instruments. One team of hardware engineers began to assess and explore the feasibility of playing different existing instruments. Categories of instruments were assessed -- brass, string, percussion, woodwind, and keyboard -- and rated based on the robot's ability to play them. Following these assessments, the hardware engineers constructed robot-friendly redesigns of a guitar and a harp for the robot to play.

In parallel, one hardware test engineer (now referred to as the lead engineer) reformulated the instrument prompt from ``What if the robots could play musical instruments?'' to ``What if the robot \textit{is} the musical instrument?'' Towards this second prompt, the hardware test engineer began analyzing signals generated by the robot to determine which signals could be involved in the novel instrument. The artist and this lead engineer met with a group of software engineers from another team that generates music using machine learning. Through these conversations, ideas formed about ways to process and analyze the signals coming off of the robot as input into a novel instrument.

Thus, the collaborative artistic and engineering team were motivated by the open-ended prompt to alter the robot's personality through musical sounds. The primary team involved in the Music Mode project were software engineers, the artist, and later, an experienced musical composer.

\subsection{Robot Setup and Existing Sounds}

The robot in use is a mobile manipulator with a 7-degree-of-freedom (DoF) arm with a 2-DoF mobile base, and a one-DoF two-finger gripper. A speaker is affixed to the front of the robot and there is a microphone on the robot's head that receives vocal commands. Each arm joint is a revolute joint and has a brushless motor to move the joint and an encoder that measures the value of the joint in space. The 2-DoF base is non holonomic, with two sets of wheels in the front and the back. There are vents with fans along the robot's arm and torso, to assist with heat dissipation. There are two brushed motors in the 2-DoF neck, with an encoder. Joint velocities (rather than positions) are used for Music Mode, and were sampled at ~250 Hz. 

The robot made a variety of physical, mechanical sounds due to the movement of its joints and its LIDAR sensor. For example, the motor in the head made a ``whirring'' noise when articulating up or down. We refer to these sounds as ``native sounds'' for the remainder of the paper, and as the ``Native'' condition in the experiments and case study. The team observed the robot performing a variety of tasks such as navigating, door opening, and table wiping, in order to gather their overall impression of the sounds.

We measured sound samples of different robot components that make noise when they move and matched these to a musical scale. The spinning sensor in the front makes a sound that is close to a B natural, the base fans at an E natural, shoulder at C sharp, gripper at a G natural, the head tilt at a C sharp, and the base wheels at an A natural. When several joints move in the arm at the same time, it sounds roughly like A3-D4 in the key of D. The Robotic condition is comprised of amplified sounds from the robot's existing system sounds. The Orchestra sound condition is a collection of classical instruments played together. The speaker at the front of the robot that plays the generated Music Mode sounds is controlled with a percent value, where is 10 percent is 65 dB and 100 percent is 80 dB.

\subsection{Implementation}

The core principle of the final Music Mode is that specific signals within the robot function as triggers for playing audio samples. These signals can be at the control level, for example the target positions of the arm joint actuators, or higher-level post-processed states like “two humans are detected in the robot’s view”. \change{The audio samples follow the paradigm of a general audio playback device. While Music Mode is enabled, any actuation activates a corresponding musical track for the duration of the movement. When the movement finishes, the musical track stops and awaits the next actuation of that joint, then the musical track restarts from the beginning.} During the development process, we first built an off-robot system to generate the sounds. This was followed by a less technically complex, but more responsive on-robot implementation.

\subsubsection{Off-robot Implementation}

\begin{figure}[ht]
\centering
\includegraphics[width=1.0\textwidth]{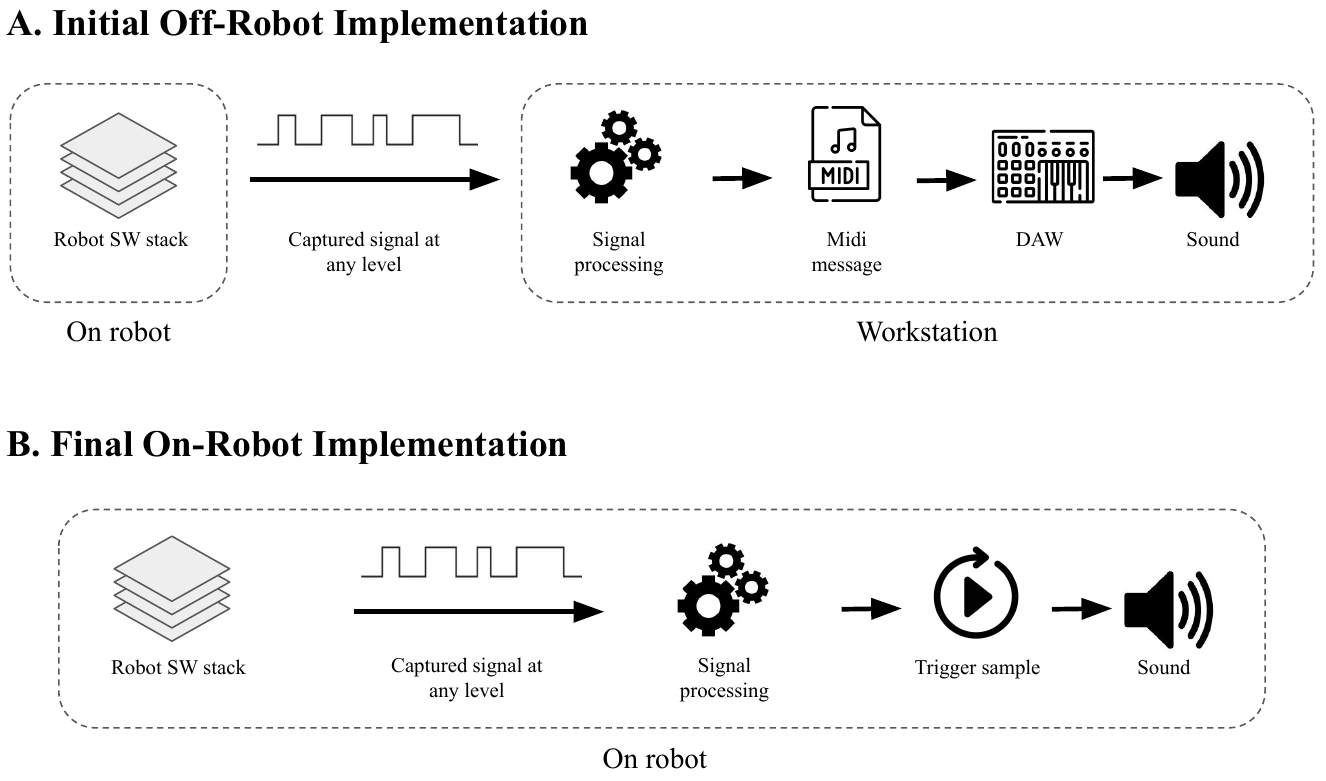}
\caption{The initial off-robot signal flow implementation (A) contrasted with the final on-robot signal flow implementation (B). A. In the off-robot signal flow, a signal is captured at any layer of the robot software stack. This signal is processed per the composer's interpretation and converted in realtime into a MIDI message on a workstation. That message is consumed by a Digital Audio Workstation (DAW) carefully tuned by the composer, and a sound is generated. This system is limited by the DAW to the processing of 16 signals at once. B. In the on-robot signal flow, a signal is captured at any layer of the robot software stack. This signal is processed, triggering a predetermined sound sample. This system can process a large quantity of signals, but provides limited options for fast creative iterations. In both implementations, the signal processing included only playing sound when the velocity was above 0.05 rad/s and the velocity was over that threshold for more than 0.04 seconds}
\label{fig:combinedimplementation}
\end{figure}

The initial experimental approach was to create the sounds mapped to the captured robot signals on a workstation connected to a speaker (Figure \ref{fig:combinedimplementation}A). This implementation offered the benefit of fast creative and technical iterations, but lacked the ability to scale the experience to many robots at once.
\change{Each robot signal was captured and processed using a Python notebook environment from a Linux computer. The processed signal was turned into a MIDI message that would serve as a trigger for a mapped sound sample at a specific pitch in Ableton Live 11, the Digital Audio Workstation (DAW), on a Mac. The sound samples were carefully crafted by a sound designer and composer to evoke a particular experience to the user.} The artist tuned the collection of sound samples, referred to as ``soundscape,'' with the following characteristics:
% It was early on established that mapping sound samples to motion was a straightforward and experiential useful approach. 

\begin{itemize}
    \item Triggering of sound samples linked to target joint velocities results in a intuitive relationship, where the visual events of an actuated joint with the auditory events are clearly coupled. Other signals introduce unnecessary noise in the captured signals (like actual joint velocities) or result in less intuitive linkage between the sound and motion (like target joint accelerations) for the user.
    \item Users can typically distinguish two to three sound samples to joint pairings synchronously before the experience becomes overbearing. Since typically not all joints move at once, we established experimentally that 5 to 7 joint to sound pairings is optimal.
    \item Joints exerting more inertia are mapped to lower pitched sounds.
    \item The soundscape consists of a diverse set of samples using sounds with different timber and theme.
    \item Samples may change and grow while played, but the thematic sensibility stays the same. 
    \item Samples can technically play in perpetuity and are designed to fade-out within a reasonable time at any given point to avoid being perceived as disruptive. 
\end{itemize}

This comprised the first Coding Prototype as shown in Figure \ref{designprocess}. Upon establishing a desired soundscape, an adapted workflow with prerecorded soundscape is implemented on the robot in order to scale the experience.

\subsubsection{On-robot Implementation}

The on-robot flow  (Figure \ref{fig:combinedimplementation}B) processes the robot signals to sounds on the robot computing system using the robot's speaker as the output. The flow enables a scalable experience by running Music Mode on any robot. The main difference from the off-robot flow is the use of localized compute resources for processing and the static processing pipeline where the robot signals trigger directly the predefined sound samples without the use of MIDI messages or a DAW. \change{The software was written in Java on a Linux machine using the Android MediaPlayer API. New builds were pushed to the robots daily, including the Music Mode software. The robot's Android computer would need to be restarted in order to get the most updated code.} 

\subsection{Audio Implementation}

\subsubsection{Early Iterations of Music Mode}
The initial music was implemented by mapping the value of joint encoders to a note on a scale, this was the first Sound Selection shown in Figure \ref{designprocess}. For example, the robot's torso joint at 80 degrees generated a middle C note on a C major scale of a digitized piano, and moving the joint to 90 degrees generated a D note on a C major scale. The artist and engineers experimented with different resolutions between degree values and their mapping to different notes on a scale (ex. 10 degrees of difference between notes versus 8 degrees of difference). In addition, they experimented with different collections of joints and scales on the same digital instrument, such as both the torso and the shoulder playing the same major scale or the torso playing a different scale than the shoulder.

These early explorations sounded strange and discordant, yet evocative. \change{They sounded discordant because they were literally playing different chords that were inharmonious together.} Because each joint was commanded separately, unusual combinations of notes and scales would be played together at any time. Due to the fact that each joint moved at a different velocity, there was no discernible time signature to the generated music, adding to the inharmonious impression of the sounds. After these initial implementation explorations with Music Mode, the artist and engineers elected to collaborate with a composer collaborator to rethink the position-to-musical-note implementation and help design the soundscape. 

\subsubsection{Orchestra Music Mode Design Development}
After this initial set of sounds, the artist and engineering team worked with a composer to create a new soundscape. The composer sent four different options for audio themes for the team to choose from. These different draft options varied in instrumentation, rhythm, and sentiment. The artist, composer, and engineer played these different sound options on their laptops while the robot was moving, to gather a sense of how the sounds' timing and overall feeling would align with the impression of the robot. After selecting one of the sound options, the composer then began to work on a finished series of sound samples, at the same time reconfiguring the Sound Selection and Final Software (Figure \ref{designprocess}). The composer strongly advocated selecting a series of instruments that would sound harmonious together, rather than a single instrument played across many robot joints. As the Music Mode sounds could be played at any time, in combination with any of the other Music Mode joint sounds, the composer was meticulous and thoughtful about the sound sample for each joint and how all the sound samples could play together in random configurations. \change{To illustrate this further, the composer elected not to use two repeating instruments on different joints, as it would not separate each joint with a distinctive sound. Further, the composer wrote specific fades for each instrument so that the appearance of the sounds would not be jarring or surprising.}

\begin{figure}[t!]
\begin{flushleft}
\includegraphics[width=.9\textwidth]{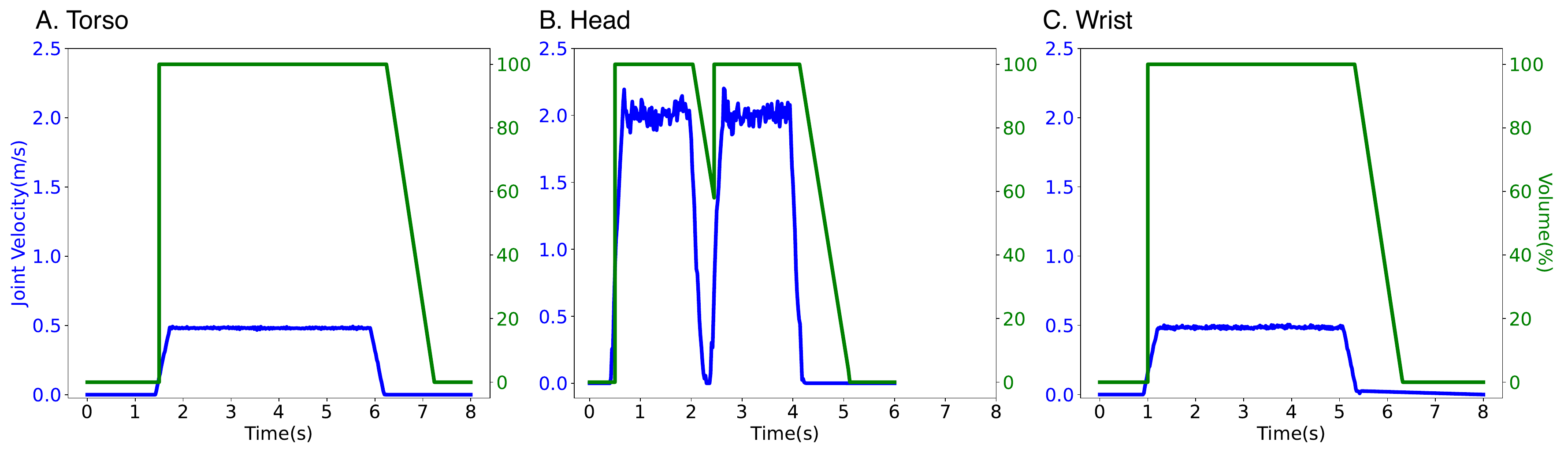}
\caption{The final on-robot sound implementation shown through joint velocities and produced sound volume. Joint velocity over time contrasted with the corresponding robot volume over time for the on-robot implementation. A. The torso joint with the fastest fade at 4\% linear fade. The percent fade means the amount the volume was reduced every 0.04 seconds. For example, a robot speaker playing at 100\% would then sound at 96\% after 0.04 seconds of the joint velocity going to zero. B. The head joint with the slowest fade at 1.5\%. C. The wrist with a fade of 3\%.}
\label{fig:vel_vol}
\end{flushleft}
\end{figure}

After the composer created a set of sounds, the full team worked together to determine which sounds would be paired with which robot joint. The team aimed to have the sounds logically pair with the joint, so the higher pitched, tinnier sounds were paired with smaller joints, while the lower pitched, rounder sounds were paired with larger joints. Pairing the joint with the sound took several days of iteration, as the composer tuned and tweaked the sounds throughout. The final Orchestra joint-to-sound pairing, as well as a short description of the sounds, is below and in Figure \ref{glamour}:
\begin{itemize}
    \item Base - Strings. When the robot's base drives around, atmospheric, light strings play.
    \item Torso - Bass. The robot's torso is a large joint, when it moves it triggers the sound of a slow, plunking bass.
    \item Shoulder - Strings. The rotating shoulder triggers a string section that is slow and melodic.
    \item Elbow - Piano. The elbow moving triggers a flittering, gentle piano sound.
    \item Hand - Bells. As the hand rotates, short bell sounds emit.
    \item Wrist - Woodwinds. The wrist rotation triggers a slow, meditative woodwind sound.
    \item Gripper - Percussion. As the robot's gripper opens and closes, a triangle makes short chimes.
    \item Head - Brass. As the head tilts or pans, a slightly staccato trumpet sound is triggered.
\end{itemize}

As each instrumental track had its own unique sound signature, it was important that each track was balanced. To achieve the desired sounds, engineers worked with the composer to add fade-outs to each track individually. These fades allowed for smoother transitions between the robot's movement and the resultant sound through the speaker. The torso, wrist, and head robot joint velocities over time are shown with the corresponding volume over time in order to show the fades and the signal response in Figure \ref{fig:vel_vol}. There are three total fades in use in Orchestra Music Mode: a 4\% linear fade, a 3\% linear fade, and a 1.5\% fade. Once the sounds were finalized, the full experience was deployed on several robots in operation (Deploy on Robot in Figure \ref{designprocess}). The Hand and Head joints were on the 1.5\% linear fade, Wrist joint on a 3\% linear fade, and all other joints on the 4\% fade.

\section{Experiment Methods} \label{Methods}

\subsection{Experiment 1 Design}
After creating Music Mode, the team wanted to understand how people perceived the robot when Music Mode was running. The team identified two research questions to pursue:
\begin{itemize}
    \item How does Music Mode affect how individuals perceive the robot's \textit{presence}?
    
    \item How does Music Mode affect how individuals perceive the robot's \textit{capabilities}?
\end{itemize}

The team designed a research study in order to examine these questions. Three sound modes were compared as independent variables in the first experiment: 
\begin{itemize}
    \item \textbf{Orchestra} - the original implementation of Music Mode, with analog orchestral instruments.
    \item \textbf{Native} - Music Mode is turned off, the only sounds heard were those generated by the robot's sensors and motors.
    \item \textbf{Robotic Sounds} - an amplification of the robot's existing sounds.
\end{itemize}
\change{The Robotic Sounds were designed by the composer and involved different variations of machine-like and existing robot motor noises. Each robot sound was again paired with a robot joint and many iterations of these pairings were tested for the Robotic condition. The composer captured the various pitches of the robot’s moving joints and parts as a starting point for the Robot Sounds. He then created the tracks from scratch and sampled aspects of motorized noises at similar pitches (ex. Gears grinding). He did not record the robot’s sounds directly. He composed several iterations when creating these sounds, similar to his artistic attention on the Orchestra Music Mode. While the Orchestra soundscape was the first iteration of Music Mode, the Robotic soundscape did receive equal care and consideration into the design of its sounds. The orchestra sounds have a more legato rhythm while the Robotic sounds have a more staccato rhythm that may be received as disjointed. This was intentional to allude to the machine nature of the robot and therefore, by design, evoke an alternate user experience.} 

\begin{figure}[t]
\centering
\includegraphics[width=\columnwidth]{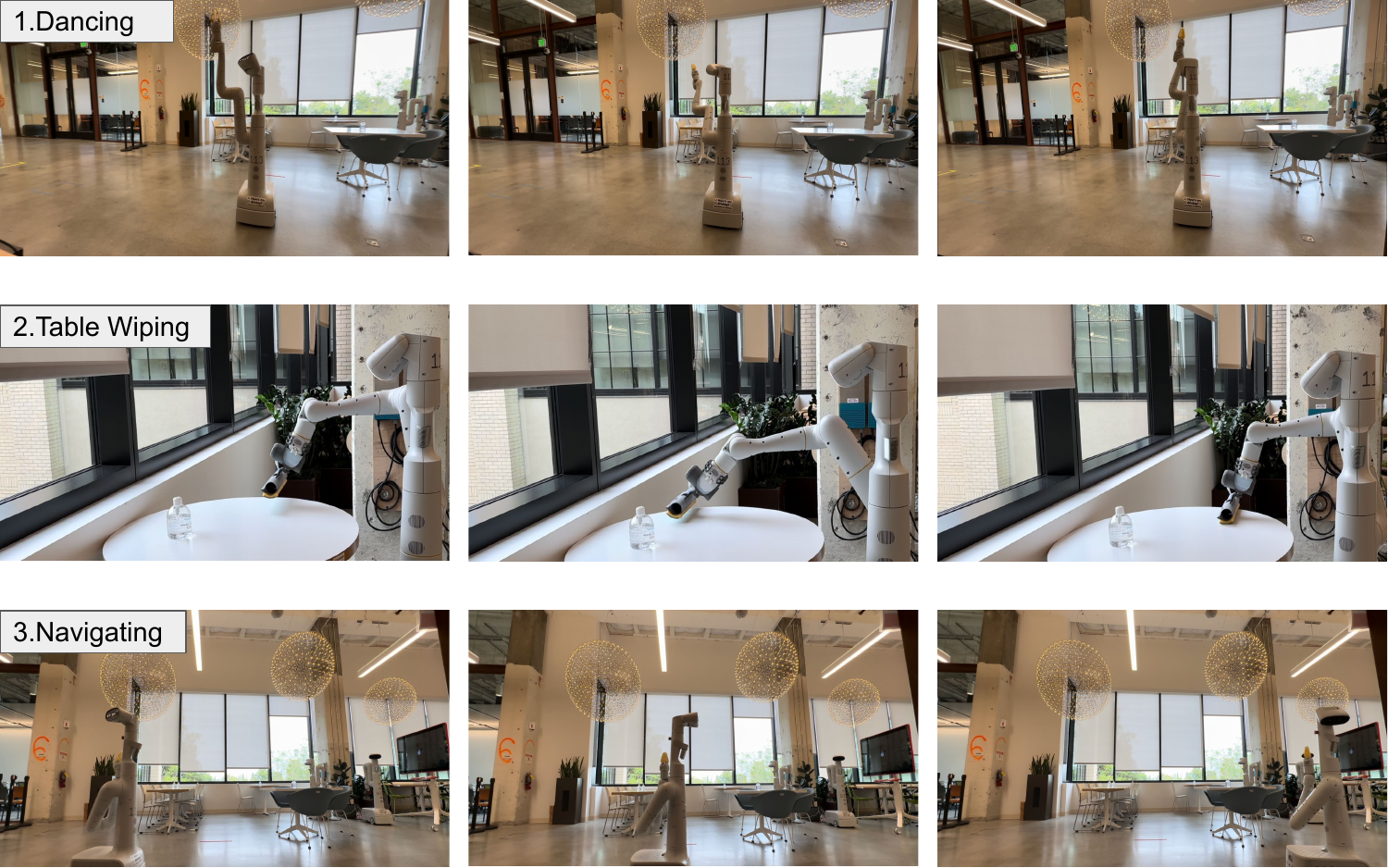}
\caption{The robot performing three different tasks during Experiment 1. The top three images show the robot performing a short choreography with its arm and head. The middle three images illustrate the robot wiping a table. The bottom three images display the robot navigating across the room. The robot performed these three tasks while running each different test condition: Orchestra, Robotic, and Native.}
\label{task_images}
\end{figure}

After these three conditions were set, there were three robot actions, shown in Figure \ref{task_images}:
\begin{itemize}
    \item \textbf{Dancing}. The robot performed a short choreographed sequence of arm and base movements.
    \item \textbf{Navigating}. The robot navigated from one side of the room to the other three times.
    \item \textbf{Table Wiping}. The robot wiped a table with a custom wiping tool attached to its gripper.
\end{itemize}

Each participant saw the actions in the same order: dancing, navigating, and table wiping, while the sound mode order was randomized. Given the three sound modes, there were six total sound condition orderings: \change{1. Orchestra, Native, Robotic Sounds, 2. Orchestra, Robotic Sounds, Native, 3. Native, Robotic Sounds, Orchestra, 4. Native, Orchestra, Robotic Sounds, 5. Robotic Sounds, Native, Orchestra, 6. Robotic Sounds, Orchestra, Native}. This was a within subjects design, so each participant saw the same independent variables. These sound modes are shown in the accompanying video: Orchestra at 0:34 seconds, Native at 1:22, and Robotic at 1:35.

The experiment lasted approximately 30 minutes, and the participant tasks were as follows:
\begin{itemize}
    \item \textbf{Intro Survey}: The participant arrived at the experiment location and filled out an introductory survey. The brief survey consisted of their contact information and background.
    \item \textbf{Observation 1}: The participant observed the robot perform all three actions (Navigating, Table Wiping, and Dancing) with the first condition of the randomized sound conditions.
    \item \textbf{Perception Survey 1}: The participant completed a survey about their perceptions of the robot.
    \item \textbf{Observation 2}: The participant observed the robot perform all three actions (Navigating, Table Wiping, and Dancing) with the second condition of the randomized sound conditions.
    \item \textbf{Perception Survey 2}: The participant completed the same survey as survey 1, this time in response to the second sound mode.
    \item \textbf{Observation 3}: The participant observed the robot perform all three actions (Navigating, Table Wiping, and Dancing) with the third condition of the randomized sound conditions.
    \item \textbf{Perception Survey 3}: The participant completed the same survey as survey 1, this time in response to the third sound mode.
    \item \textbf{Final Survey}: The participant completed a final survey.
\end{itemize}

The Perception Surveys 1, 2, and 3 were identical and included 28 total questions. The first 23 questions were the semantic differential scales ``Godspeed'' questions from \citet{bartneck2009measurement}, as shown in Figures \ref{6_axbar} and \ref{followupax}. These questions were answered on a scale from 1 to 5, where 1 indicated the person's perception of the robot aligned with the first attribute while 5 indicated the person's perception of the robot aligned with the second attribute. For example, if the pairing was ``Fake or Natural'', then a score of 1 corresponded to ``Fake'', while 5 corresponded to ``Natural''. Five additional statement questions were created, where 1 indicated "Strongly Disagree" and 5 strongly "Strongly Agree":
\begin{itemize}
    \item I liked the sounds produced by the robot.
    \item I would feel comfortable around the robot in this encounter.
    \item The robot's motion was satisfactory.
    \item It was clear what the robot wanted to do.
    \item The robot successfully performed the task.
\end{itemize}

\change{In Experiment 1, the participant was approximately 10 feet from the robot and sitting stationary at one table in the room. There were no other sounds in the room as it was a quiet corner of the office (there was minor ambient noise down the hall, since the experiment space was not fully enclosed behind a door). The only laptop in use was for the participant to complete the surveys, hence there were no other visible speakers present inside the space. As the robot was moving and turning, the sound would shift (not via the software, but as the Doppler effect similar to a firetruck going by), thus underscoring the sound source as the robot’s speaker. For example, as the robot drove across the room, it felt louder to the participant when it was nearby, and quieter when it was farther away.}

All individuals who participated in the experiment were recruited via a company-wide email list. 18 individuals participated in the controlled experiment and participation was completely voluntary. One question in the Final Survey asked about their music experience and was rated 1-5, 5 being Strongly Agree and 1 being Strongly Disagree: ``I consider myself musically inclined'' with an average of 4.11. Another question asked about their hearing, ``I would consider my hearing:'' as 1 through 5 where 5 was ``Good'' and 1 was ``Poor'', the average score for this across all participants was 4.61. 50\% of participants were Male, 38.9\% Female, and 11.1\% Non-binary/prefer not to say. 83.3\% of participants noted that they worked in a STEM field, while 16.7\% reported they worked in a non-STEM field. 33.3\% had worked with this robot previously.
Participants' answers about their race were 44.4\% White, 33.3\% Asian, 11.1\% Black, 5.6\% Middle Eastern or North African, and 5.6\% Prefer not to answer.

\subsection{Experiment 2 Design}

After completing Experiment 1, the team wanted to study a potential confounding variable: that changes in robot perception due to the Orchestra Music Mode were a result of beautiful, enjoyable music being played near the robot, rather than the music being a result of the robot's movement. To study this, the team designed a follow-up second experiment.

The study structure was as follows: participants viewed three videos of the robots (one video of each task) completing the same tasks as in Experiment 1, with the correct Orchestra Music Mode playing. After this, they completed the same condition survey as in Experiment 1. Participants then viewed the same three videos, but the sound was replaced with a random collection of Orchestra Music Mode sounds (without any relationship to the movement). \change{In the video recordings, the sounds were also played from the robot's speaker.} The participants were randomized into two groups: one group saw the correct Orchestra Music Mode videos first and the random ones second, the other group saw the random Orchestra Music Mode videos first and the correct Orchestra Music Mode videos second. 

The inclusion criterion was individuals who participated in Experiment 1, and the survey was sent via email and chat; participation was completely voluntary. One individual had left the organization and an additional six declined to complete the survey, so the total number of participants was 11. Of this group, 45\% reported they identified as Female, while 45\% identified as Male and 10\% as Non-binary. Participants' answers about their race were 54.5\% White, 27.2\% Asian, 9.1\% Black, and 9.1\% Middle Eastern or North African. 91\% of participants noted they worked in a STEM field and 45\% noted they had worked with the robot previously in their work.

\change{Both Experiments 1 and 2 included the use of the Godspeed questionnaire \citep{bartneck2009measurement}, which is one of the most widely used measure in human-robot interaction research \citep{weiss2015meta}. In \citep{carpinella2017robotic}, an exploratory factor analysis was applied to the Godspeed dimensions, and Animacy and Anthropomorphism were found to load onto each other. Other scale modes of evaluating robots have been proposed and used by researchers \citep{carpinella2017robotic, ho2010revisiting, kamide2012new}. In our work, we elected to use the Godspeed questionnaire as all five dimensions were of interest, especially the dimensions Likeability, Perceived Intelligence, and Perceived Safety. We supplemented the Godspeed questionnaire with five additional statement questions to further understand participants' perceptions of the robot.}

\subsection{Case Study Design}
\change{Based on the controlled experiment findings and the positive feedback from various teams, the Orchestra Music Mode software was deployed to every robot of this type as an easter egg feature. The maximum volume was 80 dB, roughly able to be heard from approximately 10 yards away. The minimum volume was 65 dB and was challenging to hear amongst the robot's native sounds. The noise level in the various rooms ranged from 55-65 dB. The other noises included common office sounds like keyboards, low volume conversations, various microkitchen sounds (water pouring, refrigerator running, cabinets opening), and people walking, for example. The rooms were medium crowded, slightly crowded, and not crowded at all, as return to office protocols varied over the case study period. As the robot worked and navigated, it could be encountered several times by the same person who provided comments. The passersby normally performed their work in the same offices and common areas, and the robot would navigate a similar route. The robot exclusively played the Orchestra profile during these events.}

\change{The robot operators turned on the Music Mode software one time per week at an interval of their choosing. The robot would then perform a task, such as wiping a table or sorting trash, while the robot was generating music. Passersby volunteered their comments on the robot's behavior while the music was playing and these comments were tracked and recorded by the robot operators.} 
\section{Results} \label{Results}

\subsection{Experiment 1 Results}

\begin{figure}[t!]
\centering
\includegraphics[width=.9\columnwidth]{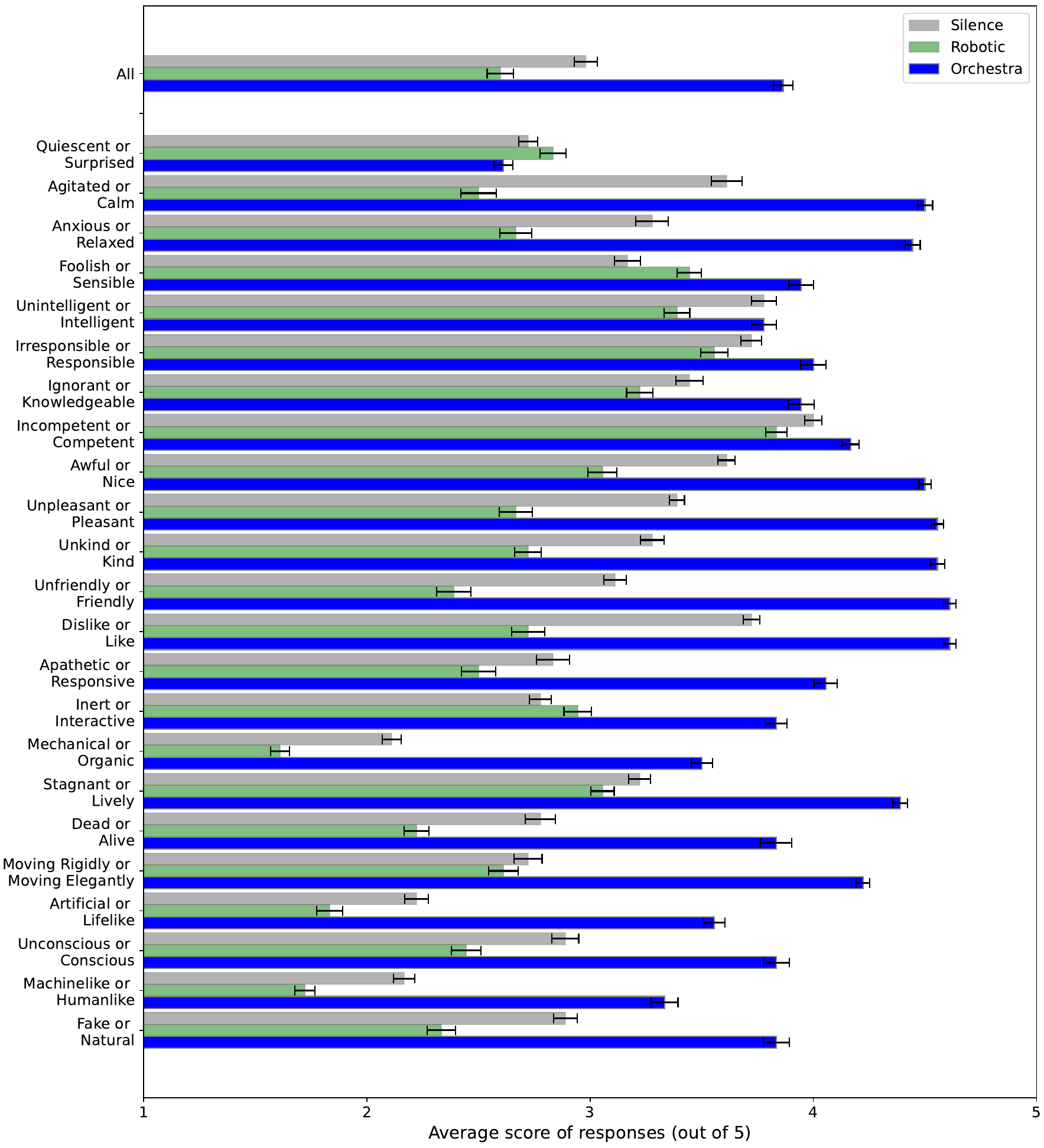}
\caption{\change{Experiment 1 results for individual Godspeed questions. Error bars indicated standard errors of the averaged score of the given question(s).}}
\label{6_axbar}
\end{figure}

\begin{figure}[t]
\centering
\includegraphics[width=\columnwidth]{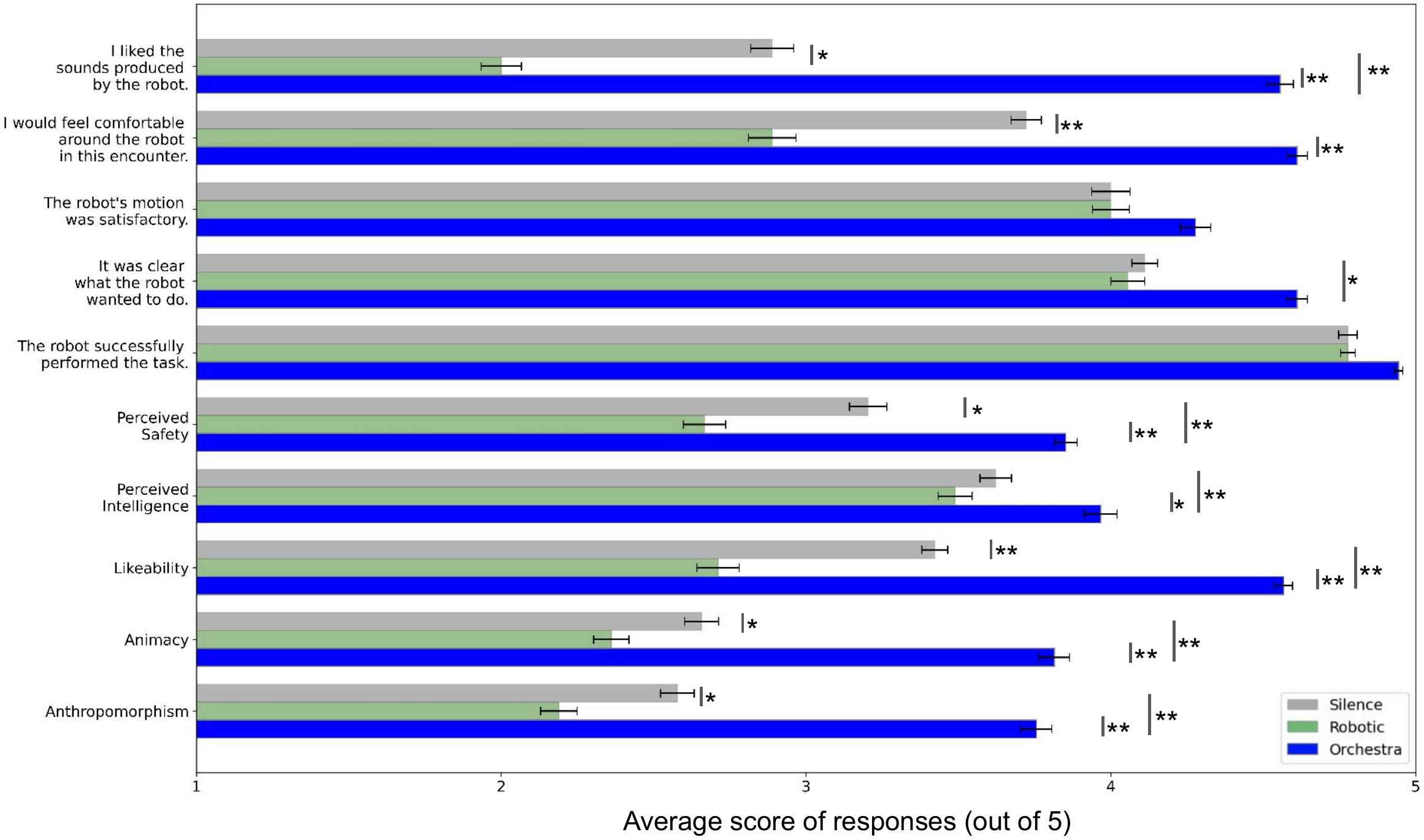}
\caption{Experiment 1 results for five statement questions and grouped averages from Godspeed categories. Error bars indicated standard errors of the averaged score of the given question(s). ``*'' indicates a statistically significant difference between two categories with p $< 0.05$, and ``**'' indicates a statistically significant difference between two categories with a p $< 0.01$.}
\label{masterbar}
\end{figure}

Responses to original Godspeed questions are shown in Figure \ref{6_axbar}. Separate repeated measures ANOVAs with Bonferroni correction were performed to compare the effect of Orchestra, Robotic, and Native conditions on each question. For the first five questions of the participant survey (I-V as described in Section \ref{Methods}), the Orchestra Music Mode condition had the highest reported score for each question. For Question II (``It was clear what the robot was going to do''), there was a statistically significant difference between the Orchestra condition and the Native condition, with a p-value $< 0.01$ and $\eta_{p}^{2}$ = 0.346. There was a statistically significant difference for Question IV, ``I would feel comfortable around the robot in this encounter'', with p-value $< 0.01$ and $\eta_{p}^{2}$ = 0.418. There was a statistically significant difference for Question V (``I liked the sounds produced by the robot''), p-value $< 0.001$ and $\eta_{p}^{2}$ = 0.500. Question III (``The robot's motion was satisfactory'') and Question I (``The robot successfully performed the task'') showed no statistically significant differences, see Figure \ref{masterbar}.

The average scores for each Godspeed question for each condition are shown in Figure \ref{masterbar}. The Orchestra condition had the highest overall average, at 3.86, while the Native condition averaged 2.97 and the Robotic at 2.59. In Bartneck \citep{bartneck2009measurement}, the individual questions A-X are grouped into five categories: Anthropomorphism: A-E, Animacy: F-K, Likeability: L-P, Perceived Intelligence: Q-U, and Perceived Safety: V-X. A repeated measures ANOVA with Bonferroni correction was performed to study the effect of each sound condition on each of the five Godspeed categories of Anthropomorphism, Likeability, Animacy, Perceived Safety, and Perceived Intelligence. For each category, considering the responses for all three tasks, we looked for statistically significant effects of sound condition with a p $< 0.01$; we found these effects statistically significant for each Godspeed category. $\eta_{p}^{2}$ values were as follows: Anthropomorphism 0.557, Animacy 0.463, Likeability 0.651, Perceived Safety 0.194, and Perceived Intelligence 0.084. \change{We did not find that the group who had previously worked with the robot had statistically significant differences in their survey responses from the group who had not worked with the robot.}

Participants were invited to write additional comments at the end of the experiment. Two participants commented on the Orchestra condition: ``[I] love the Orchestra Music Mode'' and ``the orchestral sounds played by the robot feel much more welcoming and the mechanical sounds produced by when the robot moves it's arms -- producing a sort of drill sounds seems a bit sharp.'' Others commented on the Robotic condition: ``On the robotic sounds version, I liked some sounds (driving) but hated others (arm) and so my opinion of it was conflicted'' and ``Robotic sounds remind me of a rainforest''. One participant noted that they would have liked to see the sounds continue without the movement: ``For both sound modes, it seemed a little odd how the sound would stop during temporary pauses in motion (e.g. between joint path segments, there were noticeable breaks in the sound). I feel the impressions from the sounds would have been stronger if they were more persistent while the robot was operating.''

\subsection{Experiment 2 Results}

\begin{figure}[t]
\centering
\includegraphics[width=.9\columnwidth]{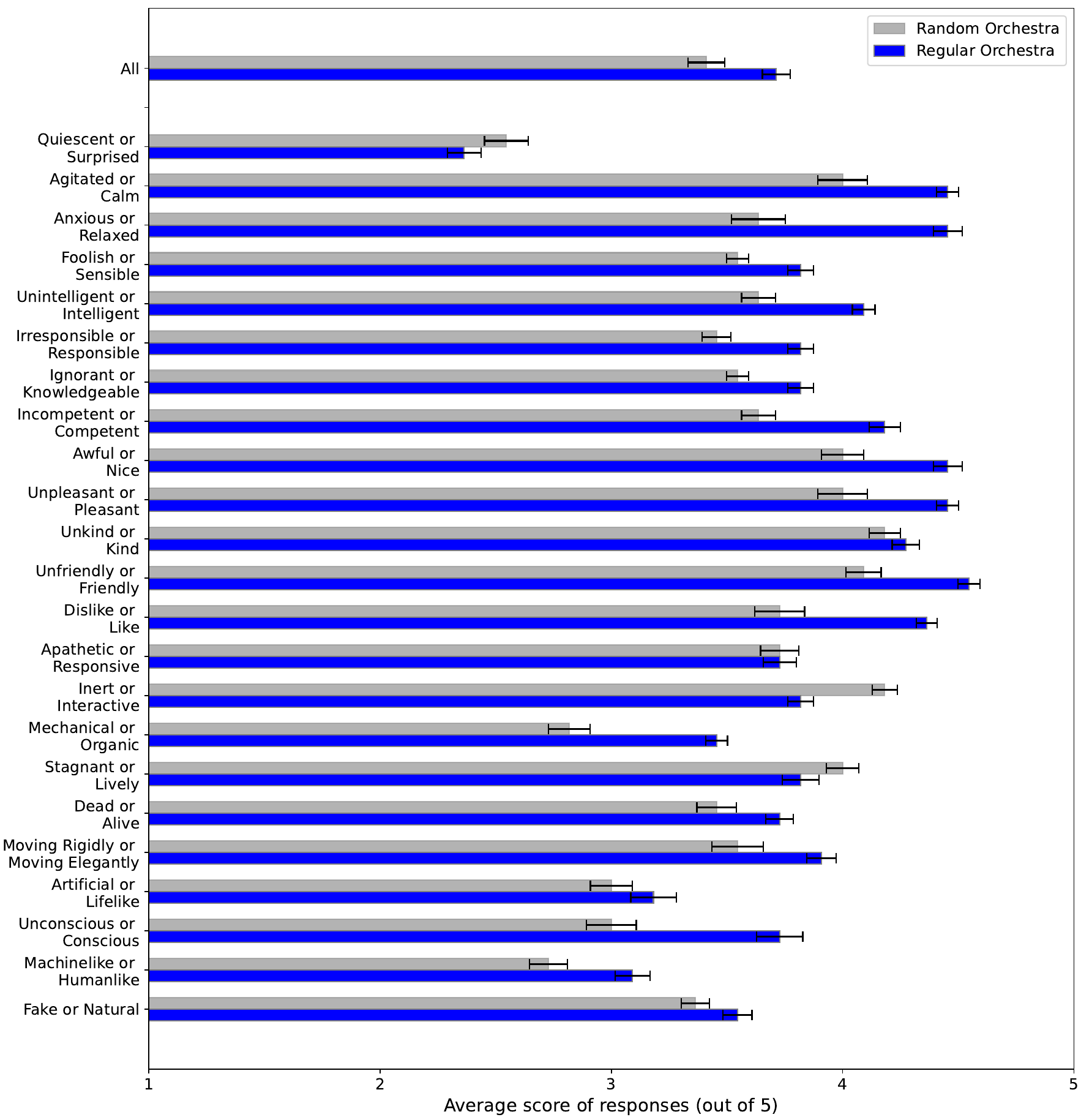}
\caption{Experiment 2 results for individual Godspeed questions. Error bars indicated standard errors of the averaged score of the given question(s).}
\label{followupax}
\end{figure}

\begin{figure}[t]
\centering
\includegraphics[width=\columnwidth]{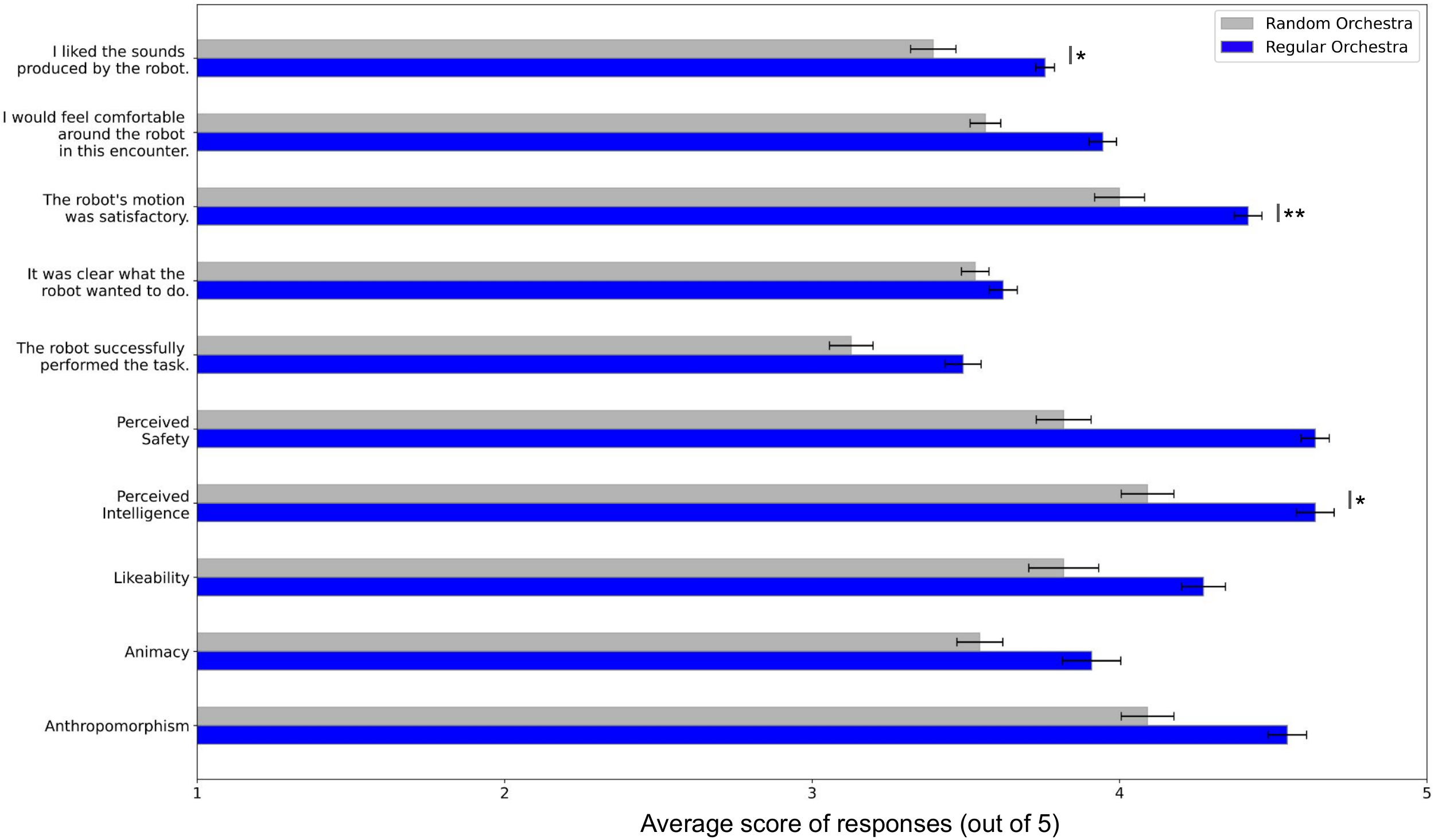}
\caption{Experiment 2 results for statement questions and Godspeed grouped categories. Error bars indicated standard errors of the averaged score of the given question(s). ``*'' indicates a statistically significant difference between the categories with p $< 0.05$, and ``**'' indicates a statistically significant difference between the categories with a p $< 0.01$.}
\label{followupbar}
\end{figure}

The 28 survey questions used in Experiment 1 were repeated for Experiment 2. The results are shown in Figure \ref{followupax} and Figure \ref{followupbar}. For the first five descriptive questions, a repeated measures ANOVA was conducted to understand the effect of the sound condition on the question rankings. ``The robot's motion was satisfactory'' was statistically significant with a p $< 0.01$ ($\eta_{p}^{2}$ = 0.582). ``I liked the sounds produced by the robot'' was statistically significant with a p $< 0.05$ ($\eta_{p}^{2}$ = 0.351). Three of the questions were not statistically significant: ``I would feel comfortable around the robots in this encounter'', ``The robot successfully performed the task'', and ``It was clear what the robot was going to do''.

Similarly, a repeated measures ANOVA was performed on the five Godspeed questionnaire categories: Anthropomorphism, Animacy, Likeability, Perceived intelligence, and Perceived safety. While the correct Orchestra Music Mode averaged higher for all five categories, the ANOVA test on Perceived intelligence was found to be statistically significant (p $<0.05$, $\eta_{p}^{2}$ = 0.343). Anthropomorphism and Likeability both approached significance while Perceived safety and Animacy were not statistically significantly different between the two conditions. \change{As in Experiment 1, we did not find that the group who had previously worked with the robot had statistically significant differences in their responses from the group who had not worked with the robot.}
%Likeability p < .15, Anthropomorphism p < .15

Participants left comments at the end of each survey. In some comments for the random Orchestra Music Mode, participants noticed the differences between the music and movement: ``It seemed as though the sounds emitted didn't always match the movements...which really affected how I saw / thought of the bot'' and ``The timing in some of the music sounded jittery, which made the robot feel jittery/glitchy to me.'' Comments for the correct Orchestra Music Mode included: ``Sounds are less jarring, the more constant and airy sounds are pleasant and kind of feel like they are guiding through the process. A bit like Teslas and other EVs that have a hum as they drive by, but more pleasant and lifelike.'' and ``The sounds produced are pleasant.''

\subsection{Case Study Results}
Comments collected during the Case Study were coded by two individuals into Positive, Neutral, and Negative; 13 comments were collected during the first month of operation. The plurality of these comments were Positive at 46.2\%, 38.5\% of comments were Neutral, and 15.4\% of comments were Negative. A selection of these comments are below: 
\begin{itemize}
    \item Positive:
    \begin{itemize}
        \item[] ``When at a good volume, it seems like a nice addition or quality of life feature for the robots. It adds onto their uniqueness.''
        \item[] ``Music Mode is nice. The music that is played distracts from the unpleasant sounds coming from the robot such as the (LIDAR) spinning and the gears moving.''
        \item[] ``I can see the music helping become a way for people to avoid bumping into the robot since there is music coming from the robot.''
        \item[] ``One person who walked by noticed the music coming from the robot, smiled, and said "That's pretty neat!''''
    \end{itemize}
    \item Neutral:
    \begin{itemize}
        \item[] ``It's strange that the music doesn't continue to play even when the robot stops. It sounds pretty static-y in that way.''
        \item[] ``Are different sounds associated with moving different joints of the robot?''
    \end{itemize}
    \item Negative
    \begin{itemize}
        \item[] "...(a) bystander the day previously...mentioned that the music was actually too loud and disturbing."
        \item[] "It's mid[dling]. I can't tell if it should be adding onto something or what the purpose of the music is for."
        \item[] "The sounds started and stopped too dramatically, would sound better if there was an ambient sound even while idling it might sound more consistent."
    \end{itemize}
\end{itemize}

These passersby comments reflect similar themes to the comments received during the controlled experiment. Themes in the Positive comments, such as the two above, included opinions that the sound emitting from the robot would assist humans in perceiving, interacting, and avoiding the robots. Neutral comments, like those above, remarked about the Music Mode software implementation and possible augmentations. Negative comments described confusion or annoyance about the sounds or volume emitting from the robot, particularly when nearby individuals were at a desk or focus area. The Music Mode software was run in a variety of different office settings, where passersby included engineers and operations staff, some of whom had seen the robots previously but others who were experiencing the robots for the first time.

General comments about the robot were also collected while the robot was not playing Music Mode. Over the same one month period as the Music Mode comments, seven comments were collected about the robot in general. Several of these comments were about the robot's capabilities and tasks, and others asked about the robot's name and spatial awareness. Five of these comments were neutral, one negative, and one positive.

\section{Discussion} \label{Discussion}

In Experiment 1 with the Orchestra, Robotic, and Native conditions, the Orchestra music condition was rated more Anthropomorphic, Likeable, and Animate and was perceived as more Intelligent and Safe (p$<$0.01 for all). The participants rated that they would feel more comfortable around the robot while it was playing the Orchestra Music Mode and that they liked the sounds better. These findings suggest that this robot could be more accepted and welcomed in the office spaces in question if it were playing the Orchestra Music Mode.

\change{The qualitative comments about the Orchestra Music Mode in Experiment 1 illustrated the participants' enjoyment of the music (ex. ``[I] love the Orchestra Music Mode''). Comments expressed mixed feelings about the Robotic Sounds music (ex. ``On the robotic sounds version, I liked some sounds (driving) but hated others (arm) and so my opinion of it was conflicted''). As there was a statistically significant difference in likeability among these conditions, how much participants liked the music may have impacted their likeability of the robot. In both the Orchestra and Robotic Sound condition modes, there was the same mapping between joint velocity and sound triggering/volume. There was not a statistically significant difference between conditions for Question III (``The robot's motion was satisfactory'') and Question I (``The robot successfully performed the task''). One participant noted, ``For both sound modes, it seemed a little odd how the sound would stop during temporary pauses in motion (e.g. between joint path segments, there were noticeable breaks in the sound). I feel the impressions from the sounds would have been stronger if they were more persistent while the robot was operating.'' This participant observation about the sound to movement mapping may be demonstrative of a broader trend that the same movement to sound mapping among conditions led to no perceived difference in the movement (Question III) or task performance (Question I).}

In Experiment 2, which included only the correct Orchestra Music Mode and the random Orchestra Music Mode conditions and was exclusively online, the correct Orchestra Music Mode was statistically significantly rated higher in terms of Perceived intelligence than the random Orchestra Music Mode. While the correct Orchestra Music Mode was higher for all other Godspeed questionnaire categories, these differences were not statistically significant. The correct Orchestra Music Mode was also rated more highly on the questions of whether the robot's movement was satisfactory and if the participant liked the sounds produced by the robot. This indicates that participants noticed a difference between the random Orchestra Music Mode and regular (movement-linked) Orchestra Music Mode, and that the matching of movement to the generated sounds leads to a more Intelligent and satisfactory movement perception. Both experiments show that positive perceptions are improved when these helper robots play the Orchestra Music Mode synchronized to movement. More work is needed to verify if simply playing beautiful music while a robot is in the environment would increase likeability on its own. In addition, Experiments 1 and 2 differed in that Experiment 2 was exclusively online and the nuanced perception of the sound being linked to movement may have be lost. 

\change{Participant comments in Experiment 2 can be linked to the quantitative results. One participant noted about the random Orchestra Music Mode: ``It seemed as though the sounds emitted didn't always match the movements...which really affected how I saw / thought of the bot''. The fact that the participant was able to notice a mismatch between the movement and sound could mean that others were able to notice it as well; this may have led to the statistically significant differences in Question III ``The robot's motion was satisfactory'' and Perceived Intelligence. Despite both conditions in Experiment 2 being comprised of the same overall soundscape, participants did describe the sounds as different between them. For the correct Orchestra Music Mode, one participant said ``Sounds are less jarring, the more constant and airy sounds are pleasant and kind of feel like they are guiding through the process. A bit like Teslas and other EVs that have a hum as they drive by, but more pleasant and lifelike'' and another noted, ``The sounds produced are pleasant''. These comments are supported by the statistically significant difference in Question I,``I liked the sounds produced by the robot''.}

The Robotic sounds and Native conditions rated lower than the Orchestra condition in all of the measured Godspeed categories for Experiment 1. The robot as a default only plays Native sounds. But, in general, it may not necessarily be inadvisable for the default robot to be rated as less Anthropomorphic and Animate. As the helper robot is aimed to be accepted and welcomed, but not necessarily social, these two categories (Anthropomorphic and Animate) may not be explicit requirements for helper robots. As the likability of the sound (Question V) was directly correlated with the robot's Godspeed likeability, future work with other robots may utilize sound as an expressive medium to alter robot perception. 

Some of these positive sentiments carried into the case study. Positive comments remarked about the robots' uniqueness while playing Music Mode and the pleasantness of the Orchestra music sounds in contrast with the robot's native sounds. The consistency across these two experiments and a case study demonstrate that the Orchestra Music Mode feature is an effective feature for improving robot acceptance and positive perceptions.

\change{The qualitative comments from both experiments and the case study have implications for future projects similar to Music Mode. Firstly, at least two participants were able to perceive that the sounds stopped when motion stopped. Another participant noted that the sounds and motions seemed to mismatch during the random Orchestra Music Mode. In later projects, the design team should ensure that the produced sound aligns with the weight or importance of a joint (for example, a high, tinny sound would seem out of place for a robot's wheeled base, as that carries the entire robot). This sound to joint mapping was an important part of the artistic and design process. Based on the goal of the sonification, researchers and industry practitioners can use these qualitative comments to alter their sound designs. If increased likability is desired, next projects should also ensure that each individual sound is pleasant, rather than only the concatenation or combination of the sounds. If future teams wanted to increase perceived intelligence and satisfaction with the robot's motion, the movement and music should be paired, rather than continuous sound.}

\change{Limitations to this work include the choice of design process: in foregrounding an interdisciplinary team and a musical series of sound samples, the team did not explore small signaling noises (such as chimes or button presses on an smartphone) or purely software generated sounds. The team used a combination of custom written survey questions and the Godspeed questionnaire, other researchers have used alternative HRI questionnaires such as RoSAS \citep{carpinella2017robotic}. It would have been ideal for all participants from Experiment 1 to participate in Experiment 2. An expanded participant pool for Experiment 2 could have led to more robust statistical outcomes.}

\change{We recognize that there is inherent subjectivity in this work, as the inclusion of an artist and composer leads to choices about sound inclusions and mappings that are not fully software generated or automatic. We posit that the fine-tuning of sound generation models and the architecting of sound generation software is a partially subjective task as the programmer/researcher makes choices about weights, inputs, outputs, and the like. In this work, the composer had 50+ years of musical expertise and the artist had 7 years of experience generating both art works and research projects with robots. We found these skills to be valuable because the design process required generating music and generating a novel robot personality/character. In order to facilitate reproducibility, we have open sourced our code as well as the sounds for this project. This work may serve as a roadmap for other interdisciplinary teams working in hybrid academic/industry settings to build and study new robot characteristics in everyday human environments.}

In future work, the effect of orchestra Music Mode may be examined under different tasks, such as when the robot is performing a regular hardware test or beginning new services. It is also worth exploring under which conditions the Orchestra Music Mode is distracting, annoying, or confusing, such as cases where people are focusing in quiet desk areas. Additional music samples, mixes, and volumes could be deployed inside the existing music framework. Another direction in future work is investigating the use of music sounds beginning \textit{before} the robot starts to move, as a pre-signaling mechanism, rather than exactly when the robot moves. \change{It would be beneficial in the future to conduct a broader case study, and examine if earlier or later exposures to the robot would alter human perceptions of them.}

Finally, as this feature emerged from a co-creation process between musicians, artists, roboticists, and engineers, the research team is compelled to embark on more interdisciplinary feature design, where such differing forms of expertise collaborate towards a shared end. In doing so, novel corners of the robotic feature design space are explored and can lead to positive outcomes. 
\section{Conclusion} \label{Conclusion}

\change{We have introduced Music Mode, a mapping between a robot's joint motions and sounds, that was created by a collaborative group of artists and engineers at a robotics company. First, related work in sound design for human-robot interaction, robots playing musical instruments, robots as musical instruments, and artists collaborating in human-robot interaction design was introduced. This work differed in that the sound conditions were the result of a collaborative artistic-engineering design process, sounds were deployed on a real robot and cued from live sensor streams, experiments were conducted while the robot was performing its regular everyday tasks, and participant pools included individuals who had previously used the robot in their work.}

\change{During the design process, the artist, engineers, and composer wrote software to stream sounds on a laptop, and then migrated this software onto the robot's native computer so the sounds could be streamed through its on-board speaker. Artistic consideration was applied to the mapping between a sound and the representative joint. An expert composer created the musical sounds that appeared in the different conditions and several iterations of each sound sample were made.}

\change{Two experiments and one case study were conducted that tested the default Orchestral Music Mode, a Robotic Sounds Music Mode, and the robot's Native sounds. In the first experiment, the Orchestral Music Mode condition led to the robots being rated as statistically significantly more likeable, anthropomorphic, animate, perceived safe, and perceived intelligent than the other two conditions. The Orchestral Music Mode was also rated as statistically significantly more likable sounds, and higher participant comfort than the other two conditions; it was statistically significantly perceived as having more clarity compared to the Native sounds condition. The case study reinforced positive perceptions of the Orchestral Music Mode as the plurality of comments were positive.}

\change{This artistic and design process lays one template for interdisciplinary teams in an academic or private institution to collaborate on human-facing robot behaviors. The results from this paper could be used in future design processes in order to modulate a robot's perceived personality in a human-facing setting, while the robot performs everyday tasks. As robots continue to enter everyday environments and complete tasks near humans, interdisciplinary projects such as these may be design and engineering guides for the complex requirements of these new domains.}

\begin{acks}
We thank Peter Van Straten, for being an essential artistic collaborator and composing all the music in this project; Adrian Li for initial software development work and creative input; and Daniel Lam for his role as an integral developer. We received technical and strategic contributions from Kyle Jeffrey, Sara Ahmadi, Denise Gamboa, and Hans Peter Brondmo. We appreciate Leila Takayama for providing guidance on survey questions, experiment design, and statistics, and Sean Follmer for his feedback and suggestion to include Experiment 2. Thank you to Steve Hoang, Nicholas Castro, Norman Chiu, Tiffany Ku, and Meghha Dhoke for recording feedback from passersby.
\end{acks}

\bibliographystyle{acm-reference-format} %IEEEtran
\bibliography{bibliography}

\appendix \label{Appendix}
% \section{Appendix Section 1}








\end{document}